\documentclass[10pt,twocolumn,letterpaper]{article}

\usepackage{cvpr}              % To produce the CAMERA-READY version
\usepackage{graphicx}
\usepackage{amsmath}
\usepackage{amssymb}
\usepackage{booktabs}
\usepackage{bigstrut}
\usepackage{multirow}
\usepackage{bm}
\usepackage{makecell}
\usepackage{color}
\usepackage[table]{xcolor}
\usepackage{booktabs}
\usepackage{graphicx}

\usepackage[labelformat=simple]{subcaption}

\usepackage[pagebackref,breaklinks,colorlinks]{hyperref}

\usepackage[capitalize]{cleveref}
\crefname{section}{Sec.}{Secs.}
\Crefname{section}{Section}{Sections}
\Crefname{table}{Table}{Tables}
\crefname{table}{Tab.}{Tabs.}

\def\cvprPaperID{5072} % *** Enter the CVPR Paper ID here
\def\confName{CVPR}
\def\confYear{2023}

\makeatletter
\def\thanks#1{\protected@xdef\@thanks{\@thanks
        \protect\footnotetext{#1}}}
\makeatother
\begin{document}

\title{Spectral Enhanced Rectangle Transformer for Hyperspectral Image Denoising}
\author {
    % Authors
    Miaoyu Li $^{1 \dagger}$, 
    Ji Liu $^{2 \dagger}$,
   Ying Fu $^{1*}$\thanks{${^\dagger}$Equal Contribution, ${^*}$Corresponding Author},
    Yulun Zhang\textsuperscript{\rm 3},
    Dejing Dou\textsuperscript{\rm 4}\\
    \textsuperscript{\rm 1}Beijing Institute of Technology, \textsuperscript{\rm 2}Baidu Inc., \textsuperscript{\rm 3}ETH Z\"{u}rich, \textsuperscript{\rm 4}BCG X \\
    {\tt\small miaoyli@bit.edu.cn, liuji04@baidu.com, fuying@bit.edu.cn},\\ {\tt\small yulun100@gmail.com, dejingdou@gmail.com}
}

% \affiliations {
%     % Affiliations
%     \textsuperscript{\rm 1} Beijing Institute of Technology,  
%     \textsuperscript{\rm 2} Baidu Inc,
%     \textsuperscript{\rm 3} ETH Z\"{u}rich. \\
%     \{miaoyli, fuying\}@bit.edu.cn, yulun100@gmail.com
% }

% \author{Miaoyu Li\\
% Beijing Institute of Technology\\
% Beijing, China\\
% {\tt\small miaoyli@bit.edu.cn}
% \and
% Ji Liu\\
% Baidu Inc.\\
% Beijing, China\\
% {\tt\small liuji04@baidu.com}
% \and
% Ying Fu\\
% Beijing Institute of Technology\\
% Beijing, China\\
% {\tt\small fuying@bit.edu.cn}
% \and
% Yulun Zhang\\
% \\
% \\
% {\tt\small yulun100@gmail.com}
% \and
% Dejing Dou\\
% BCG X\\
% Beijing, China\\
% {\tt\small dejingdou@gmail.com}
% }
\maketitle

%%%%%%%%% ABSTRACT
\begin{abstract}
   Denoising is a crucial step for hyperspectral image (HSI) applications. Though witnessing the great power of deep learning, existing HSI denoising methods suffer from limitations in capturing the non-local self-similarity. Transformers have shown potential in capturing long-range dependencies, but few attempts have been made with specifically designed Transformer to model the spatial and spectral correlation in HSIs. In this paper, we address these issues by proposing a spectral enhanced rectangle Transformer, driving it to explore the non-local spatial similarity and global spectral low-rank property of HSIs. For the former, we exploit the rectangle self-attention horizontally and vertically to capture the non-local similarity in the spatial domain. For the latter, we design a spectral enhancement module that is capable
of extracting global underlying low-rank property of spatial-spectral cubes to suppress noise, while enabling the interactions among non-overlapping spatial rectangles. Extensive experiments have been conducted on both synthetic noisy HSIs and real noisy HSIs, showing the effectiveness of our proposed method in terms of both objective metric and subjective visual quality. The code is available at \textcolor{magenta}{https://github.com/MyuLi/SERT}.% \textcolor{red}{low-rank}
\end{abstract}

\section{Introduction}
\label{sec:intro}

With sufficient spectral information, hyperspectral images (HSIs) can provide more detailed characteristics to distinguish from different materials compared to RGB images. Thus, HSIs have been widely applied to face recognition~\cite{uzair2015hyperspectral,uzair2013hyperspectral}, vegetation detection~\cite{burai2015classification}, medical diagnosis~\cite{wei2019medical}, \emph{etc}. 
With scanning designs~\cite{basedow1995hydice} and massive wavebands, the photon numbers in individual bands are limited. HSI is easily degraded by various noise. 
Apart from poor visual effects, such undesired degradation also negatively affects the downstream applications. To obtain better visual effects and performance in HSI vision tasks, denoising is a fundamental step for HSI analysis and processing.

Similar to RGB images, HSIs have self-similarity in the spatial domain, suggesting that similar pixels can be grouped and denoised together. Moreover, since hyperspectral imaging systems are able to acquire images at a nominal spectral resolution, HSIs have inner correlations in the spectral domain. Thus, it is important to consider both spatial and spectral domains when designing denoising methods for HSI.
Traditional model-based HSI denoising methods~\cite{chen2010denoising,fu2015adaptive,he2015total} employ handcrafted priors to explore the spatial and spectral correlations by iteratively solving the optimization
problem. Among these works, total variation \cite{he2015total,he2018hyperspectral,zhang2019hyperspectral} prior, non-local similarity \cite{he2019non}, low-rank~\cite{chang2017hyper,chang2020weighted} property, and sparsity~\cite{wei2017structured} regularization are frequently utilized. The performance of these methods relies on the accuracy of handcrafted priors. In practical HSI denoising, model-based methods are generally time-consuming and have limited generalization ability in diverse scenarios. 

To obtain robust learning for noise removal, deep learning methods~\cite{yuan2018hyperspectral,sidorov2019deep,wei20203,cao2021deep} are applied to HSI denoising and achieve impressive restoration performance. However, most of these works utilize convolutional neural networks for feature extraction and depend on local filter response to separate noise and signal in a limited receptive field. 

Recently, vision Transformers have emerged with competitive results in both high-level tasks~\cite{wang2021pyramid,dosovitskiy2020image} and low-level tasks~\cite{zamir2022restormer,bandara2022hypertransformer,chu2021twins}, showing the strong capability of modeling long-range dependencies in image regions. To diminish  the unaffordable quadratically computation cost to image size, many works have investigated the efficient design of spatial attention\cite{chen2022activating,ye2021perceiving,yang2022scalablevit}. Swin Transformer~\cite{liu2021swin} splitted feature maps into shifted
square windows. CSWin Transformer \cite{dong2022cswin} developed a stripe window across the features maps to enlarge the attention area. As HSI usually has large feature maps, exploring the similarity beyond the noisy pixel can cause unnecessary calculation burden. Thus, how to efficiently model the non-local spatial similarity is still challenging for HSI denoising Transformer.

HSIs usually lie in a spectral low-rank subspace ~\cite{chang2017hyper}, which can maintain the distinguished information and suppress noise. This indicates that the non-local spatial similarity and low-rank spectral statistics should be jointly unitized for HSI denoising. However, existing HSI denoising methods~\cite{huang2018joint,xiong2021mac} mainly utilize the low-rank characteristics through matrix factorization, which is based on a single HSI and requires a long-time to solve. The global low-rank property in large datasets is hardly considered.

In this paper, we propose a \textbf{S}pectral \textbf{E}nhanced \textbf{R}ectangle \textbf{T}ransformerc (SERT) for HSI denoising. To reinforce model capacity with reasonable cost, we develop a multi-shape rectangle self-attention module to comprehensively explore the non-local spatial similarity. Besides, we aggregate the most informative spectral statistics to suppress noise in our spectral enhancement module, which projects the spatial-spectral cubes into low-rank vectors with the assistance of a global spectral memory unit. The spectral enhancement module also provides interactions between the non-overlapping spatial rectangles. With our proposed Transformer, the spatial non-local similarity and global spectral low-rank properly are jointly considered to benefit the denoising process. Experimental results show that
our method significantly outperforms the state-of-the-art methods in both simulated data and real noisy HSIs.
	
Overall, our contributions can be summarized as follows:
 \begin{itemize}
 \vspace{-2mm}
 	\setlength{\itemsep}{0.5mm}
 	\setlength{\parsep}{0pt}
 	\setlength{\parskip}{0pt}
	\item We propose a spectral enhanced rectangle Transformer for HSI denoising, which can well exploit both the non-local spatial similarity and global spectral low-rank property of noisy images.
	\item We present a multi-shape rectangle spatial self-attention module to effectively explore the comprehensive spatial self-similarity in HSI.
	\item A spectral enhancement module with memory blocks is employed to extract the informative low-rank vectors from HSI cube patches and suppress the noise.

\end{itemize}
\section{Related Works}
\subsection{Hyperspectral Image Denoising}
%As a fundamental pre-processing for HSI applications, 
HSI denoising is a well-developed research area in computer vision \cite{chang2017hyper,he2019non,xiong2022smds} and remote sensing \cite{yuan2018hyperspectral,shi2021hyperspectral}. Mainstream HSI denoising methods can be classified into model-based methods and deep learning methods.

Traditional model-based methods~\cite{chen2010denoising,yuan2012hyperspectral,lu2015spectral,lu2015spectral,zheng2018hyperspectral} illustrate noise removal as an iterative optimization problem with handcrafted priors.
Adaptive spatial-spectral dictionary methods are proposed in \cite{fu2015adaptive}. Chang \etal \cite{chang2017hyper} employed the hyper-Laplacian regularized unidirectional low-rank tensor recovery method to utilize the structure correlation in HSI. The spatial non-local similarity and global spectral low-rank
property are integrated in \cite{he2019non} for denoising.  
Besides, other conventional
spatial regularizers \cite{lu2015spectral,zhang2019hyperspectral} and low-rank regularization~\cite{chang2020weighted} are also introduced to model
the spatial and spectral properties of noisy HSI.

%low-rank non-local prior sparse
 With great potential to automatically learn and represent features, deep learning methods~\cite{wei20203,cao2021deep,xiong2021mac,pan2022sqad} have been actively investigated for HSI denoising. Spectral-spatial features are exploited via residual convolutional network in HSID-CNN \cite{yuan2018hyperspectral}. A deep spatial-spectral global reasoning network is proposed in ~\cite{cao2021deep} to consider both the local and global information for HSI denoising. Besides, a quasi-recurrent neural network was extended to HSI denoising task~\cite{wei20203,pan2022sqad}, showing the benefits of both convolutional and recurrent neural networks. Model-guided interpretable networks have also been actively explored in~\cite{bodrito2021trainable,xiong2022smds}. Different from those convolution-based networks that have limited receptive field and fixed feature extraction paradigms, our proposed method utilizes a transformer to better model the inner similarity in spatial and spectral domains.

\subsection{Vision Transformer}
\noindent\textbf{Transformer for RGB images.}
Transformers have been actively applied to vision tasks~\cite{dosovitskiy2020image,wang2021pyramid,ye2021perceiving,fu2022low} due to its powerful ability in modeling long-range dependencies. Self-attention mechanism has been proven to be efficacious in previous works~\cite{wang2018non,hu2018squeeze}. When applied to the spatial region, it is crucial for the Transformers to consider the trade-off between computation cost and model capacity. To cut down the quadratic computation growth to image size, Dosovitskiy \etal ~\cite{dosovitskiy2020image} first employed Transformer for image recognition with images spitted in small patches. Swin Transformer~\cite{liu2021swin} was proposed with shifted window for self-attention in the spatial domain. To further enlarge the receptive field of self-attention, down-sampled attention was introduced in ~\cite{chu2021twins,wang2021pyramid,ye2021perceiving}. Without spatial information loss, Dong \etal \cite{dong2022cswin} employed horizontal and vertical stripes to compute self-attention. However, for HSI denoising, the non-local spatial similarity is not efficiently explored as these Transformers conducted the spatial self-attention in limited windows or introduced unnecessary computation cost. Besides, the combined consideration of the spatial and spectral domains are rarely investigated. 
\begin{figure*}[t]
	\scriptsize 
	\centering	
	\includegraphics[width=0.9\linewidth]{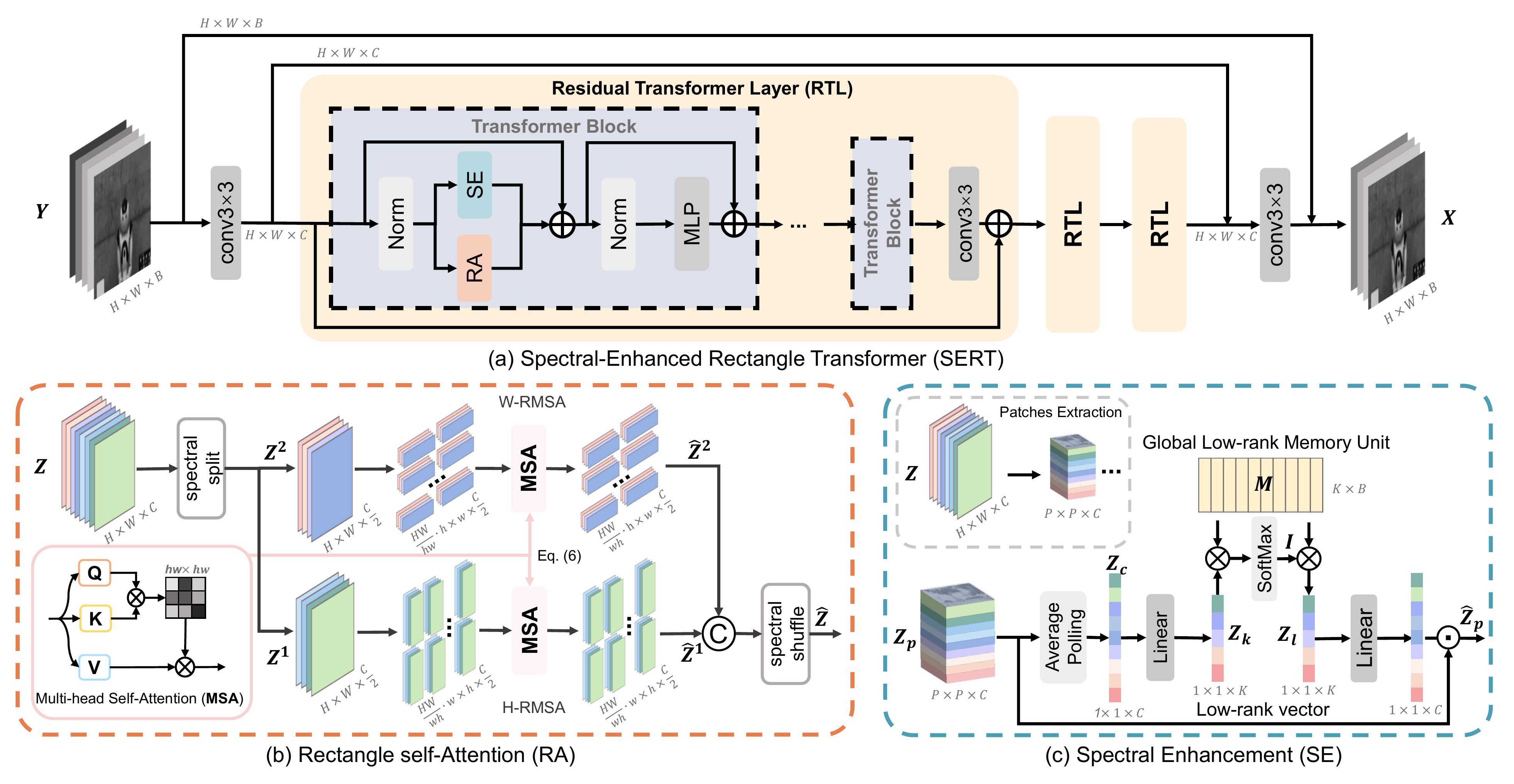}             \vspace{-3mm}                   
	\caption{Overall framework of SERT. (a) SERT mainly includes two essential components, \emph{i.e.}, SE for non-local spatial similarity and SE for global low-rank property. (b) spatial rectangle self-attention (RA) and (c) spectral enhancement (SE) module. }
	
	\label{fig:overall}
	\vspace{-2mm}
\end{figure*}

\noindent\textbf{Transformer for HSI.} Recently, there is an emerging trend of using Transformer to HSI restoration~\cite{bandara2022hypertransformer,zhang2022multiscale,su2022transformer} and HSI classification~\cite{hong2021spectralformer,liu2022dss}. An architecture
search framework was proposed in ~\cite{zhong2021spectral} to find a suitable network consisting of spectral and spatial Transformer for HSI classification. A 3D quasi-recurrent and Transformer network was presented in~\cite{bandara2022hypertransformer} for hyperspectral image denoising, which combined the 3D quasi-recurrent layer with Swin blocks. 
Different from these works that tend to directly employ existing transformer blocks to another tasks, methods in \cite{cai2022mask,cai2022coarse} solve the HSI reconstruction problem with task-oriented transformer block under the guidance of degradation mask. 
However, these works do not consider the similarity in both spatial and spectral domains.
%Without constraints such as degradation matrix in other restoration tasks,
Here, we introduce our spectral enhanced rectangle Transformer to HSI denoising, exploring the most important two characteristics of HSI, including spatial non-local similarity and global low-rank properties.  

%-------------------------------------------------------------------------
\section{Spectral Enhanced Rectangle Transformer}
Assuming the degraded noisy HSI as $\bm{Y} \in \mathbb{R}^{H\times W\times B}$, where $H$, $W$, and $B$ represent the height, width, and band of the HSI, the noise  degradation can be formulated as
\begin{equation}
\bm{Y} = \bm{X} + \bm{n},
\end{equation}
where $\bm{X}$$\in$$\mathbb{R}^{H\times W\times B}$ is the desired clean HSI, and $\bm{n}$$\in$$\mathbb{R}^{H\times W\times B}$ denotes the addictive random noise. In realistic HSI degradation situations, HSIs are corrupted by various types of noise, \emph{e.g.}, Gaussian noise, stripe noise, deadline noise, impulse noise, or a mixture of them.

In this section, we elaborately introduce our proposed spectral enhanced rectangle Transformer for HSI denoising. 
The overall architecture is shown in Figure \ref{fig:overall}. In our implementation, each Residual Transformer Layer (RTL) consists of 6 Transformer blocks. And the proposed Transformer Block mainly contains two essential components, \emph{i.e.},  rectangle self-attention (RA) module and spectral enhancement (SE) module. Figure \ref{fig:overall}\textcolor{red}{(b)} and Figure \ref{fig:overall}\textcolor{red}{(c)} illustrate the detailed framework of RA module and SE module, respectively.  The outputs of RA and SE are added together to achieve comprehensive feature embeddings for noise removal. Next, we discuss each module in detail.

\subsection{Spatial Rectangle Self-Attention}
To remove noise from HSI, it is important to explore the similarity information in spatial domain~\cite{he2019non}, which implies that similar pixels can be aggregated together for denoising. Existing deep learning-based HSI denoising methods mainly utilize the convolutional layer to extract the local information with spatially invariant kernels, limiting the flexibility to model the non-local similarity.

\begin{figure}[t]
	\centering	
	\scriptsize
		\setlength{\tabcolsep}{1.42mm}
			\begin{tabular}{cccccc}
		\multicolumn{6}{c}{\includegraphics[width=0.98\columnwidth]{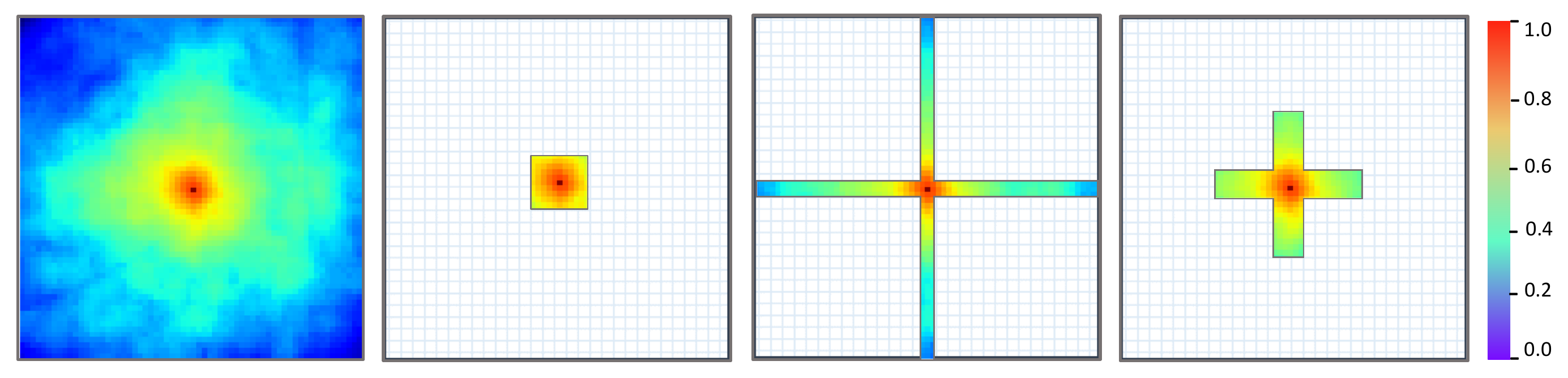}	}
 \\
	 &\makecell[c]{(a) similarity\\ to central pixel} &   \makecell[c]{\quad (b) window\\ \quad self-attention}  & \makecell[c]{(c) cross-shaped \\ self-attention}&  \makecell[c]{\hspace{-3mm}(d) rectangle \\ self-attention (ours)}

	\end{tabular}
	
	\vspace{-3.5mm}
	\caption{ This similarity statistic is obtained via Realistic dataset\cite{zhang2019hyperspectral}. As the distance becomes longer, the similarity decreases.}
	\label{fig:heatmap}
    \vspace{-4mm}
\end{figure}

For better model capacity, there are various attempts~\cite{zamir2022restormer,wang2021pyramid,liu2021swin} that employ Transformer as an alternative solution to convolution neural network. The power of self-attention mechanism in modeling spatial information has also been proven in~\cite{liang2021swinir,chu2021twins}.
Since the global self-attention in the spatial domain introduces high computational complexity, Swin Transformer~\cite{liu2021swin} and CSWin Transformer \cite{dong2022cswin} split the input feature into windows or stripes for attention operation. 
From the heatmap shown in Figure \ref{fig:heatmap}, we can observe that neighboring pixels are more similar to the center pixel than distant pixels. When conducting spatial self-attention, Swin (see Figure \ref{fig:heatmap}\textcolor{red}{(b)}) focuses on local information while CSwin (Figure \ref{fig:heatmap} \textcolor{red}{(c)}) tends to utilize pixels which is less informative. Thus, how to effectively conduct the self-attention in the informative spatial regions to model non-local similarity is still challenging for HSI denoising.

Here, we propose a rectangle self-attention in the spatial domain, in which the feature maps are split into several non-overlapping rectangles. As shown in Figure \ref{fig:heatmap}, our rectangle Transformer focuses on the informative neighboring pixels and obtains more exhaustive information in non-local area. At different stages of the network, rectangles of different shapes are employed to explore better expression ability. 

The details of our proposed RA module are shown in Figure \ref{fig:overall}\textcolor{red}{(b)}. To obtain comprehensive features, the rectangle self-attention is conducted in vertically and horizontally after the spectral split operation. Different from \cite{dong2022cswin}, we add a spectral shuffle \cite{ma2018shufflenet} operation to exchange the information from two branches. Since rectangle self-attention in vertical and horizontal focuses on different regions and has different receptive fields, the shuffle operation also enlarges the respective field of the whole module.

Let $\rm \bm{Z}$$\in$$\mathbb{R}^{H\times{W}\times{C}}$ denote the input features of RA module. The outputs of RA module is calculated via
\begin{align}
	&\bm{Z}_1,\bm{Z}_2  = {\rm{Split}}(\bm{Z}), \\
	&\hat{\bm{Z}}^1 = {\rm{W\text{-}RMSA}}(\bm{Z}^1),\hat{\bm{Z}}^2 = {\rm{H\text{-}RMSA}}(\bm{Z}^2)\\
	& \hat{\bm{Z}}  = {\rm{Shuffle}}([\hat{\bm{Z}}^1,\hat{\bm{Z}}^2]),
\end{align}
where $\rm{W\text{-}RMSA}$ denotes the horizontal rectangle multi-head self-attention, and $\rm{H\text{-}RMSA}$  denotes the vertical rectangle multi-head self-attention. $\bm{Z}$ is firstly divided into two parts in spectral domain, where $\bm{Z}^1$$\in$$\mathbb{R}^{H\times{W}\times{\frac{C}{2}}}$ and $\bm{Z}^2$$\in$$\mathbb{R}^{H\times{W}\times{\frac{C}{2}}} $. Then, $\bm{Z}^1$ and $\bm{Z}^2$ conduct the $\rm{W\text{-}RMSA}$ and  $\rm{H\text{-}RMSA}$ separately.

Supposing the size of horizontal rectangle as [$h$, $w$] and $h$$>$$w$, for $\rm{W\text{-}RMSA}$, the input features $\bm{Z}^1$ is partitioned into non-overlapping rectangles as  $\{\bm{Z}_1^{1},\bm{Z}_2^{1},...,\bm{Z}_N^{1}\}$, in which $Z_i^{1}\in \mathbb{R}^{h\times{w}\times\frac{C}{2}}$ and $N$=$\frac{W\times H}{h\times w}$. The output of each rectangle from $\rm{W\text{-}RMSA}$ is calculated as
\begin{align}
	&\bm{Q}_{i}^1 =\bm{Z}_{i}^1{\bm{W}_q^1},\quad  \bm{K}_{i}^1 =\bm{Z}_{i}^1{\bm{W}_k^1},\quad  \bm{V}_{i}^1 =\bm{Z}_{i}^1{\bm{W}_v^1} \\
	&\hat{\bm{Z}_{i}^1} = {\rm SoftMax}({\bm{Q}_{i}^1}{\bm{K}_{i}^1}^T /\sqrt{d} + \bm{P})\bm{V}_{i}^{1},\label{eq:qkv}
\end{align}
where $\bm{W}_{q}^1$, $\bm{W}_{k}^1$, $\bm{W}_{v}^1$$\in$$\mathbb{R}^{\frac{C}{2}\times\frac{C}{2}}$ are the projection mappings of \emph{query} $\bm{Q}_{i}^1$$\in$$\mathbb{R}^{h\times{w}\times\frac{C}{2}}$, \emph{keys} $\bm{K}_{i}^1$$\in$$\mathbb{R}^{h\times{w}\times\frac{C}{2}}$, and \emph{value} $\bm{V}_{i}^1$$\in$$\mathbb{R}^{h\times{w}\times\frac{C}{2}}$.    $\bm{P}$ is the learnable parameter embedding the position and $d$ is the feature dimension. Then the outputs of horizontal rectangle self-attention is aggregated by
\begin{equation}
    {\rm{W\text{-}RMSA}}(\bm{Z}^1) = {\rm Merge}(\hat{\bm{Z}_{1}^1},\hat{\bm{Z}_{2}^1}, ..., \hat{\bm{Z}_{N}^1}).
\end{equation}

For vertical rectangle self-attention $\rm{H\text{-}RMSA}$, the size of the rectangle is [$w$, $h$] while other operations are similar to $\rm{W\text{-}RMSA}$. Moreover, at different layers of the network, rectangles in various shapes are employed to explore non-local similarity in different scales.

\subsection{Spectral Enhancement}

%\subsubsection{spectral fusion}

In traditional model-based HSI denoising methods, HSI is always represented by its extracted patches, and the low-rank property is widely explored in HSI  denoising~\cite{chang2017hyper}, compressive sensing~\cite{dong2014compressive}, unmixing~\cite{huang2018joint}, implying that the low-dimensional spectral subspace is beneficial to HSI tasks. 
We also adopt the low-rank property to guide the HSI denoising process. However, without strong regularization like SVD decomposition~\cite{chang2020weighted}, projecting the noisy HSI into a proper subspace is difficult. Thus, instead of introducing orthogonal linear projection as in~\cite{cheng2021nbnet} to HSI, we use the memory unit (MU) to store the low-rank statistics of HSI cubes. The network itself automatically learns how to represent the HSI cubes in subspace. The MU module can
be denoted as a dictionary of global low-rank spectral vectors.

As shown in Figure \ref{fig:overall}\textcolor{red}{(c)}, the features are firstly partitioned into several cube patches of size $P\times$$P$$\times{C}$ to explore the spectral-spatial correlation. In the implementation, $P$ is set to the long side of the rectangle in RA module. Accordingly, the spectral enhancement block also provides information interactions between the inside rectangles. Moreover, shift operation~\cite{liu2021swin} is employed in spatial domain to establish connections between adjacent cube patches. 

The input of SE module is denoted as  $\bm{Z}_{p}\in\mathbb{R}^{P\times{P}\times{C}}$. To obtain distinguished spectral information in a subspace, following \cite{hu2018squeeze} and \cite{chen2022activating}, a squeeze operation is employed and aggregates the features across the cube patch $\bm{Z}_{p}$ to produce a projected spectral vector of size $1\times$$1$$\times{K}$. Specifically, a downsample operation is firstly conducted in the spatial domain to obtain aggregated spectral vector $\bm{Z}_c$$\in$${\mathbb{R}^{1\times{1}\times{C}}}$. Then, it is projected to obtain  $\bm{Z}_k\in{\mathbb{R}^{1\times{1}\times{K}}}$, which is in a  subspace of rank $K$. The extraction is described as
\begin{align}
	&\bm{Z}_c = {\rm AveragePool}(\bm{Z_p}), \\
	&  \bm{Z}_k = \bm{Z}_c{\bm{W}_k},
\end{align}
where $\bm{W}_k\in\mathbb{R}^{C\times{K}}$ is the projection mapping. Notably, instead of conducting a global aggregation on the whole image, we focus on the information inside the cube since neighboring pixels tend to share similar spectral statistics.

%-------------------------------------------------------------------------

\begin{table*}[htbp]
	\centering
	\scriptsize
	\resizebox{\textwidth}{!}{
		\setlength{\tabcolsep}{0.8mm}
		\renewcommand{\arraystretch}{0.65}
		\begin{tabular}{c|ccc|ccc|ccc|ccc|ccc}
			\bottomrule[0.8pt]
		 \rowcolor[rgb]{ .949,  .949,  .949}  & \multicolumn{3}{c|}{10} & \multicolumn{3}{c|}{30} & \multicolumn{3}{c|}{50} & \multicolumn{3}{c|}{70} & \multicolumn{3}{c}{10-70} \bigstrut\\
		 	 \rowcolor[rgb]{ .949,  .949,  .949} 	\multirow{-2}[4]{*}{Method}   & PSNR  & SSIM  & SAM   & PSNR  & SSIM  & SAM   & PSNR  & SSIM  & SAM   & PSNR  & SSIM  & SAM   & PSNR  & SSIM  & SAM \bigstrut[t]\\
			\toprule[0.8pt]
		%	\hline
    Noisy & 28.13 & 0.8792 & 18.72 & 18.59 & 0.5523 & 37.9  & 14.15 & 0.3476 & 49.01 & 11.23 & 0.2301 & 56.45 & 17.24 & 0.4782 & 41.94 \bigstrut\\
    BM4D~\cite{maggioni2012nonlocal}  & 40.78 & 0.9930 & 2.99  & 37.69 & 0.9872 & 5.02  & 34.96 & 0.9850 & 6.81  & 33.15 & 0.9554 & 8.40  & 36.62 & 0.9770 & 5.51 \bigstrut\\
    LLRT~\cite{chang2017hyper}   & 46.72 & 0.9983 & 1.60  & 41.12 & 0.9920 & 2.52  & 38.24 & 0.9830 & 3.47  & 36.23 & 0.9732 & 4.46  & 40.06 & 0.9860 & 3.24 \bigstrut\\
    NGMeet~\cite{he2019non} & \textbf{47.90} & 0.9988 & 1.39  & 42.44 & 0.9816 & 2.06  & 39.69 & 0.9658 & 2.49  & 38.05 & 0.9531 & 2.83  & 41.67 & 0.9937 & 2.19 \bigstrut\\
     HSID-CNN~\cite{yuan2018hyperspectral} & 43.14 & 0.9918 & 2.12  & 40.30 & 0.9854 & 3.14  & 37.72 & 0.9746 & 4.27  & 34.95 & 0.9521 & 5.84  & 39.04 & 0.9776 & 3.71 \bigstrut\\
    
     GRNet~\cite{cao2021deep} & 45.25 & 0.9976 & 1.83  & 42.09 & 0.9957 & 2.18  & 40.25 & 0.9936 & 2.42  & 38.95 & 0.9914 & 2.63  & 41.44 & 0.9944 & 2.27 \bigstrut\\
    QRNN3D~\cite{wei20203} & 45.61 & 0.9977 & 1.80  & 42.18 & 0.9955 & 2.21  & 40.05 & 0.9929 & 2.63  & 38.09 & 0.9883 & 3.42  & 41.34 & 0.9938 & 2.42 \bigstrut\\
    T3SC~\cite{bodrito2021trainable}  & 45.81 & 0.9979 & 2.02  & 42.44 & 0.9957 & 2.44  & 40.39 & 0.9933 & 2.85  & 38.80 & 0.9904 & 3.26  & 41.64 & 0.9942 & 2.61 \bigstrut\\
    MAC-Net~\cite{xiong2021mac} & 45.20 & 0.9974 & 1.87  & 42.10 & 0.9955 & 2.35  & 40.09 & 0.9931 & 2.79  & 38.64 & 0.9905 & 3.16  & 41.31 & 0.9941 & 2.52 \bigstrut\\

        \textbf{SERT (Ours)}  & 47.72 & \textbf{0.9988} & \textbf{1.36} & \textbf{43.56} & \textbf{0.9969} & \textbf{1.77} & \textbf{41.33} & \textbf{0.9949} & \textbf{2.05} & \textbf{39.82} & \textbf{0.9929} & \textbf{2.30} & \textbf{42.82} & \textbf{0.9957} & \textbf{1.88} \bigstrut\\
    \toprule[0.8pt]
    \end{tabular}%
	}
	\vspace{-3.5mm}
	\caption{Averaged results of different methods under Gaussian noise levels on ICVL dataset. PSNR is in dB.}
	\label{tab:10_70_icvl}%
	\vspace{-2mm}
\end{table*}%

\begin{table*}[htbp]
	\centering
	\scriptsize
	\resizebox{\textwidth}{!}{
		\setlength{\tabcolsep}{0.8mm}
		\renewcommand{\arraystretch}{0.65}
    \begin{tabular}{c|ccc|ccc|ccc|ccc|ccc}
     \bottomrule[0.8pt]
    \rowcolor[rgb]{ .949,  .949,  .949}  & \multicolumn{3}{c|}{Non-i.i.d Gaussian} & \multicolumn{3}{c|}{Gaussian+Deadline} & \multicolumn{3}{c|}{Gaussian+Impulse} & \multicolumn{3}{c|}{Gaussian+Stripe} & \multicolumn{3}{c}{Gaussian+Mixture} \bigstrut\\
 \rowcolor[rgb]{ .949,  .949,  .949}   \multirow{-2}[4]{*}{Method}        & PSNR  & SSIM  & SAM   & PSNR  & SSIM  & SAM   & PSNR  & SSIM  & SAM   & PSNR  & SSIM  & SAM   & PSNR  & SSIM  & SAM \bigstrut\\
    \toprule[0.8pt]
    Noisy & 18.29 & 0.5116 & 46.20 & 17.50 & 0.4770 & 47.55 & 14.93 & 0.3758 & 46.98 & 17.51 & 0.4867 & 46.98 & 13.91 & 0.3396 & 51.53 \bigstrut\\
    BM4D~\cite{maggioni2012nonlocal}  & 36.18 & 0.9767 & 5.78  & 33.77 & 0.9615 & 6.85  & 29.79 & 0.8613 & 21.59 & 35.63 & 0.9730 & 6.26  & 28.01 & 0.8419 & 23.59 \bigstrut\\
    LLRT~\cite{chang2017hyper}  & 34.18 & 0.9618 & 4.88  & 32.98 & 0.9559 & 5.29  & 28.85 & 0.8819 & 18.17 & 34.27 & 0.9628 & 4.93  & 28.06 & 0.8697 & 19.37 \bigstrut\\
    NGMeet~\cite{he2019non} & 34.90 & 0.9745 & 5.37  & 33.41 & 0.9665 & 6.55  & 27.02 & 0.7884 & 31.20 & 34.88 & 0.9665 & 5.42  & 26.13 & 0.7796 & 31.89 \bigstrut\\
    HSID-CNN~\cite{yuan2018hyperspectral}  & 39.28 & 0.9819 & 3.80  & 38.33 & 0.9783 & 3.99  & 36.21 & 0.9663 & 5.48  & 38.09 & 0.9765 & 4.59  & 35.30 & 0.9588 & 6.29 \bigstrut\\
    GRNet~\cite{cao2021deep} & 35.19 & 0.9780 & 5.19  & 33.78 & 0.9744 & 5.42  & 32.78 & 0.9606 & 8.26  & 34.85 & 0.9772 & 5.41  & 30.91 & 0.9617 & 8.26 \bigstrut\\
    QRNN3D~\cite{wei20203} & 42.18 & 0.9950 & 2.84  & 41.69 & 0.9942 & 2.61  & 40.32 & 0.9914 & 4.31  & 41.68 & 0.9943 & 2.97  & 39.08 & 0.9892 & 4.80 \bigstrut\\
    T3SC~\cite{bodrito2021trainable} & 41.95 & 0.9922 & 4.18  & 39.59 & 0.9924  & 4.86  & 37.85 & 0.9843 & 6.53  & 41.32 & 0.9937 & 3.27  & 35.53 & 0.9767 & 8.12 \bigstrut\\
    MAC-Net~\cite{xiong2021mac}   & 39.98 & 2.9662 & 4.55  & 36.68 & 0.9860 & 5.63  & 34.54 & 0.9553 & 10.20 & 39.03 & 0.9910 & 4.03  & 30.59 & 0.9300 & 14.51 \bigstrut\\
     \textbf{SERT (Ours)}  & \textbf{44.20} & \textbf{0.9971} & \textbf{1.69} & \textbf{43.66} & \textbf{0.9969} & \textbf{1.99} & \textbf{42.67} & \textbf{0.9959} & \textbf{2.30} & \textbf{43.68} & \textbf{0.9969} & \textbf{1.97} & \textbf{40.00} & \textbf{0.9937} & \textbf{2.84} \bigstrut\\
   \toprule[0.8pt]
    \end{tabular}%

	}
	\vspace{-3.5mm}
	\caption{Averaged results of different methods under complex noise on ICVL dataset. PSNR is in dB.}
	\vspace{-3mm}
	\label{tab:complex_icvl}%
\end{table*}%
To explore the spatial-spectral correlation beyond the current HSI cube and enhance the expression ability of low-rank spectral vector, we introduce a memorizing unit (MU) to store the spectral information. 
The MU module maintains a global memory bank
$\bm{M}\in\mathbb{R}^{K\times{B}}$, which is learned as parameters of the network. For spectral vector $Z_k$, we seek the most relevant spectral low-rank vectors in MU and use these vectors to assist in adjusting the projected vector  $\bm{Z}_k$. The corresponding coefficients $\bm{I}\in\mathbb{R}^{1\times{B}}$ between $\bm{Z}_k$ and stored low-rank vectors $\bm{M}$ is  extracted by
\begin{align}
	\bm{I} = {\rm Softmax}(\bm{Z}_k{\bm{M}}).
\end{align}

With coefficients matrix $\bm{I}$, the desired low-rank vector $\bm{Z}_{l}$$\in$${\mathbb{R}^{1\times{1}\times{K}}}$ can be  obtained from MU via
\begin{align}
	\bm{Z}_{l}=\bm{I}\bm{M}.
\end{align}

Since $\bm{Z}_l$ represents the most informative spectral statistics of the noisy cube, to enhance the spatial-spectral correlation and suppress noise, we use the obtained low-rank vector as guidance to benefit the denoising process. The output of our spectral enhancement module is obtained by rescaling the input SHI cube $\bm{Z}_p$ with $\bm{Z}_l$ as
\begin{equation}
	\hat{\bm{Z}_p} = \bm{Z}_p \cdot \bm{W}_c{\bm{Z}_l},
\end{equation}
where $\bm{W}_c\in{\mathbb{R}^{C\times{K}}}$ is the project mapping and $\cdot$ is the element-wise dot product. 

\section{Experiments}
In this section, we first evaluate our method with synthetic experiments, including Gaussian noise cases and complex noise cases. Then we report
results on real noisy datasets. Finally, we perform
model analysis experiments to verify the effectiveness of the proposed
model.
			
We compare several traditional model-based HSI denoising methods including the filter-based method (BM4D~\cite{maggioni2012nonlocal}), tensor-based method (LLRT~\cite{chang2017hyper}), and orthogonal
basis-based method (NGMeet~\cite{he2019non}). Five state-of-the-art deep learning-based methods, \emph{i.e.},
HSID-CNN~\cite{yuan2012hyperspectral}, GRNet~\cite{cao2021deep}, QRNN3D~\cite{wei20203}, T3SC~\cite{bodrito2021trainable}, and MAC-Net~\cite{cao2021deep} are also compared. Traditional methods are
programmed in Matlab with Intel Core i9-10850K CPU. Our
 method as well as other deep networks is
evaluated with an NVIDIA RTX 3090 GPU. 
Peak signal-to-noise ratio (PSNR), structural similarity index metric
(SSIM) and spectral angle mapper (SAM) are used as the quantitative criteria.

\begin{figure*}
	\centering
	\scriptsize
	\setlength{\tabcolsep}{0.01cm}
	\renewcommand{\arraystretch}{0.8}
	\begin{tabular}{ccccccccccc}
		\rotatebox{90}{\tiny{nachal$\_$0823$-$1038}} &
 \includegraphics[width=1.9cm]{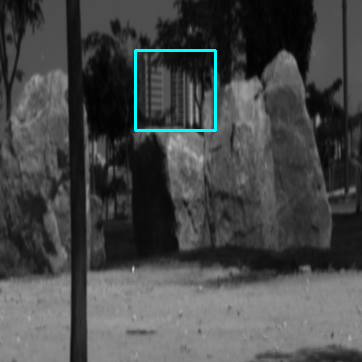}
					& \includegraphics[width=1.9cm]{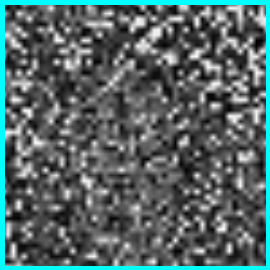}
		& \includegraphics[width=1.9cm]{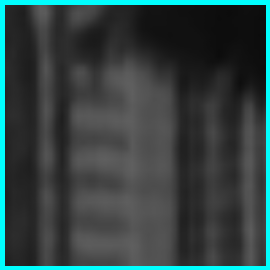}
		& \includegraphics[width=1.9cm]{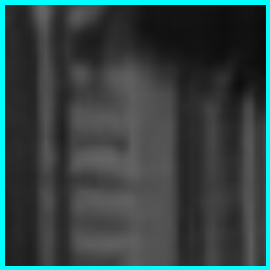}
		&\includegraphics[width=1.9cm]{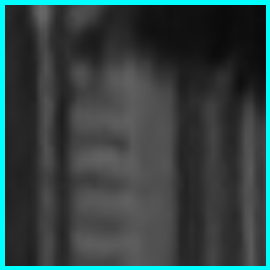}
		&\includegraphics[width=1.9cm]{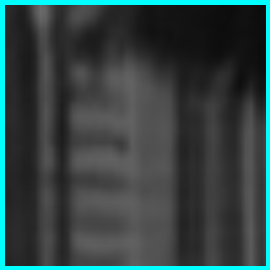}
		& \includegraphics[width=1.9cm]{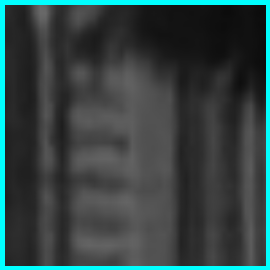}
		& \includegraphics[width=1.9cm]{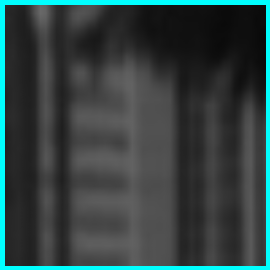}
		& \includegraphics[width=1.9cm]{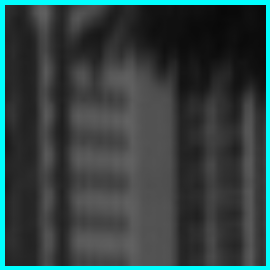}
	\bigstrut\\
		
		\rotatebox{90}{\tiny{gavyam$\_$0823$-$0933}} &
 \includegraphics[width=1.9cm]{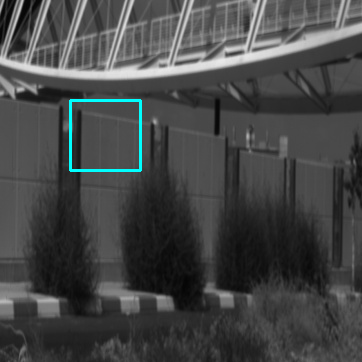}

			&	\includegraphics[width=1.9cm]{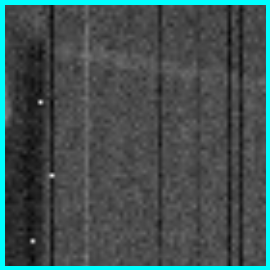}
		& \includegraphics[width=1.9cm]{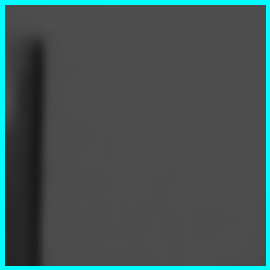}
		& \includegraphics[width=1.9cm]{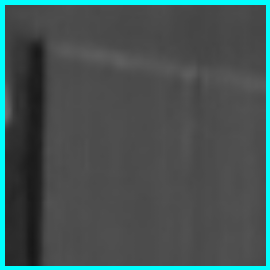}

		&\includegraphics[width=1.9cm]{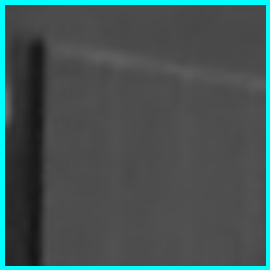}
		&\includegraphics[width=1.9cm]{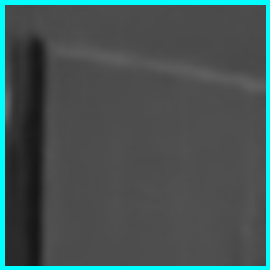}
		& \includegraphics[width=1.9cm]{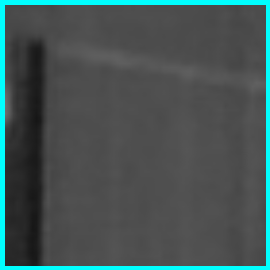}
		& \includegraphics[width=1.9cm]{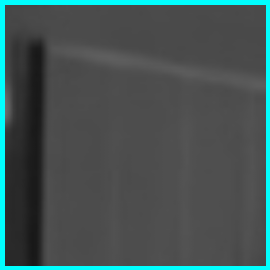}
		& \includegraphics[width=1.9cm]{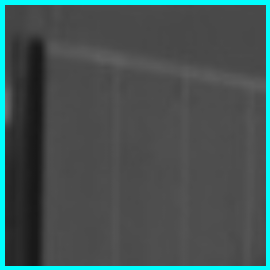}
	\bigstrut\\
		& Clean	  & Noisy & NGMeet &  HSID-CNN& QRNN3D&  T3SC & MAC-Net& \textbf{SERT (Ours)} & GroundTruth\bigstrut\\

	\end{tabular}
	\vspace{-3.5mm}
	\caption{Visual comparison on ICVL. Images are from band 28. The top row exhibits the results under Gaussian noise with noise level 50 and the bottom row exhibits the results under deadline noise.  }
	\vspace{-2mm}
	\label{fig:de_icvl} 
\end{figure*}

\begin{figure*}
 \centering
 	\scriptsize
 \setlength{\tabcolsep}{0.05cm}

  \begin{tabular}{ccccc}
  \includegraphics[width=3.2cm]{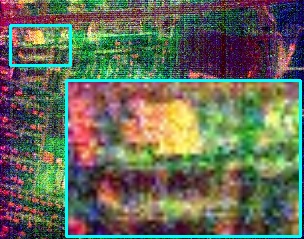}
  & 	 \includegraphics[width=3.2cm]{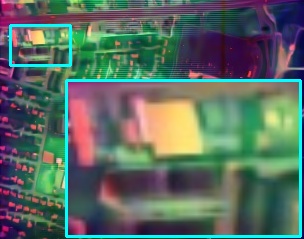}
		& \includegraphics[width=3.2cm]{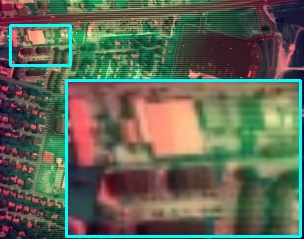}
		& \includegraphics[width=3.2cm]{imgs/urban/NGMeetcolor.jpg}
		& \includegraphics[width=3.2cm]{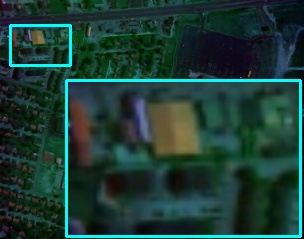}\\
   
  Noisy&  BM4D & LLRT& NGMeet &  HSID-CNN  \\ 

  	\includegraphics[width=3.2cm]{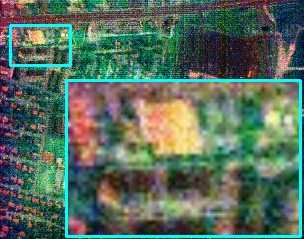}
	&	\includegraphics[width=3.2cm]{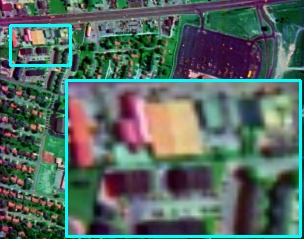}
		&\includegraphics[width=3.2cm]{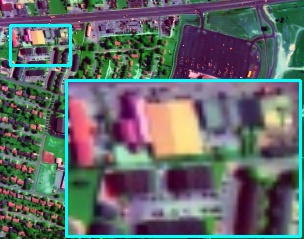}
		& \includegraphics[width=3.2cm]{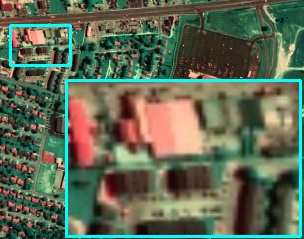}
		& \includegraphics[width=3.2cm]{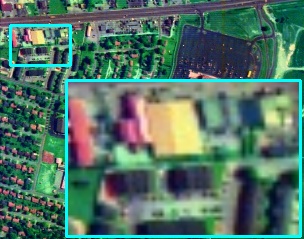}\\
  GRNet&QRNN3D&  T3SC & MAC-Net &  \textbf{SERT (Ours)} \\
  \end{tabular}

	\vspace{-4mm}
	\caption{Visual quality comparison of real noisy HSI experiments on Urban dataset with bands 1, 108, 208.}
    \vspace{-3mm}
	\label{fig:urban_dataset} 
\end{figure*}

\subsection{ Experiments on Synthetic Data}
\noindent \textbf{Datasets.} Synthetic experiments are conducted on ICVL dataset, which has been widely used for simulated studies~\cite{bodrito2021trainable,wei20203}. ICVL contains 201 HSIs of size 1392$\times$1300 with 31 bands from 400 $nm$ to 700 $nm$. We use 100 HSIs for training, 5 HSIs for validating, and 50 HSIs used for testing. Following  settings in ~\cite{bodrito2021trainable} and \cite{wei20203}, training images are cropped to size 64$\times$64 at different scales. During the testing phase, HSIs are cropped to 512$\times$512$\times$31 to obtain an affordable computation cost for traditional methods.

\begin{figure*}
	\centering
	\scriptsize
	\setlength{\tabcolsep}{0.05cm}
	\begin{tabular}{ccccc}
		
		\includegraphics[width=3.2cm]{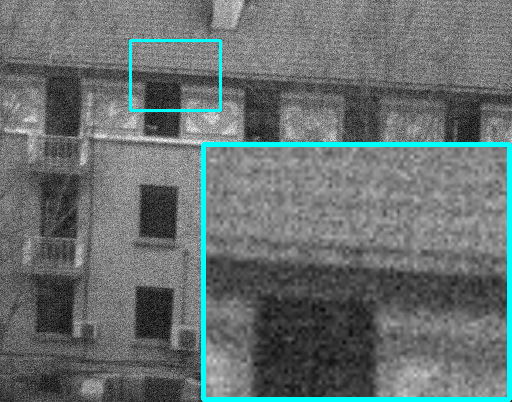}
		& \includegraphics[width=3.2cm]{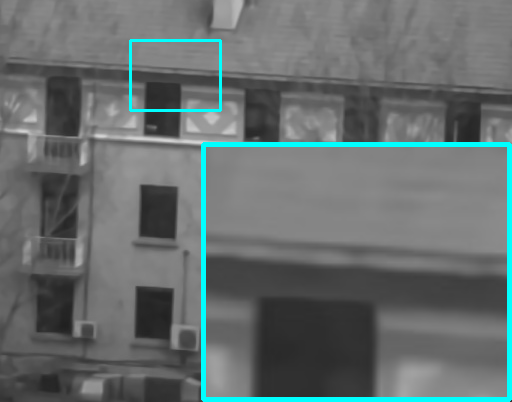}
		
		& \includegraphics[width=3.2cm]{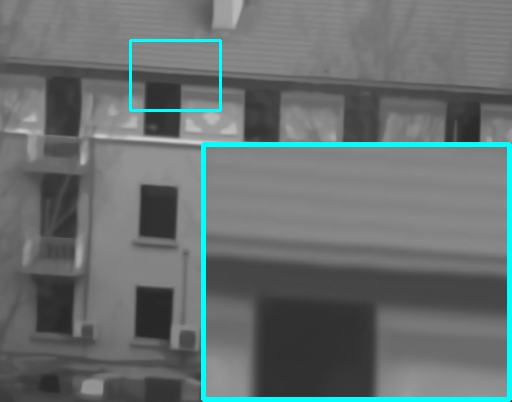}
		& \includegraphics[width=3.2cm]{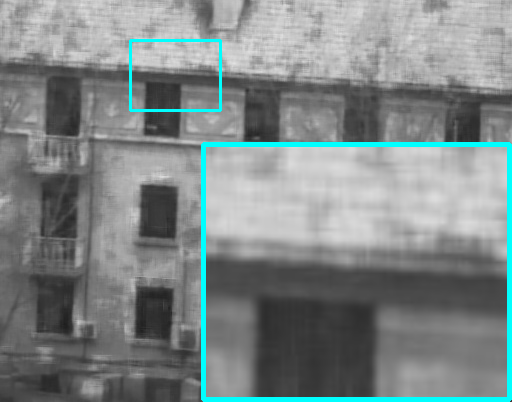}
		&		\includegraphics[width=3.2cm]{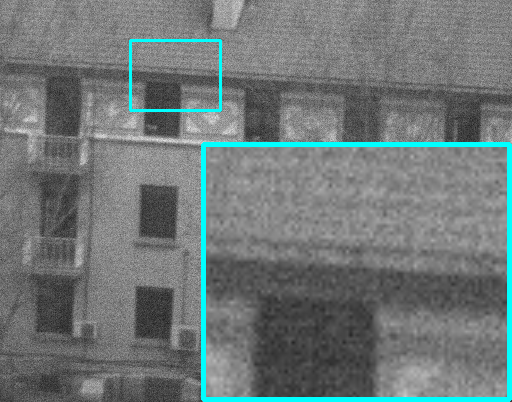}

		\\
		Noisy, 20.14 &  BM4D, 23.83 & NGMeet, 22.72 &  HSID-CNN, 22.19 &GRNet, 23.62  \\

		\includegraphics[width=3.2cm]{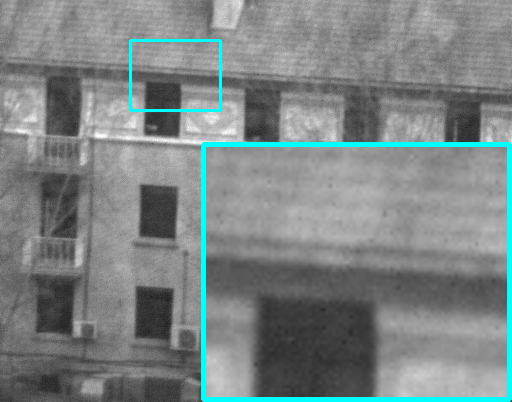}
		& \includegraphics[width=3.2cm]{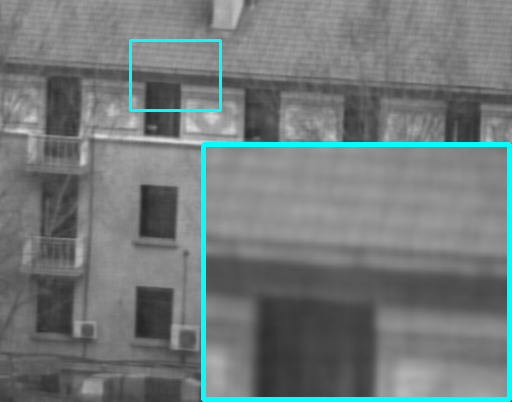}
		
		& \includegraphics[width=3.2cm]{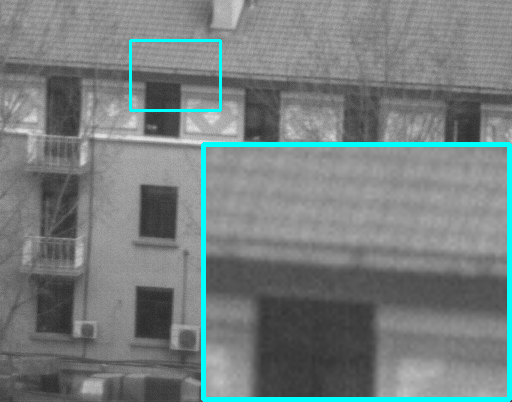}
		& \includegraphics[width=3.2cm]{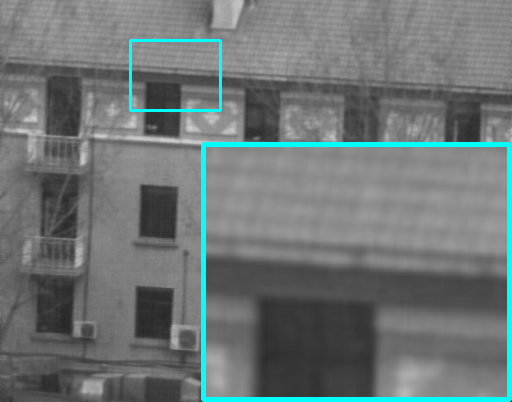}
		&\includegraphics[width=3.2cm]{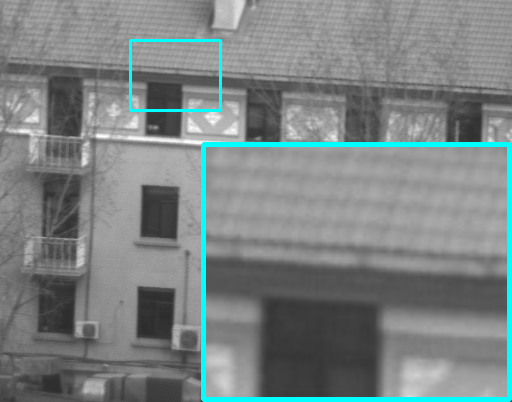}\\
		QRNN3D, 27.26&  T3SC, 26.20 &MAC-Net, 27.57  &  \textbf{SERT (Ours)}, \textbf{27.78} & GroundTruth, PSNR (dB)\\
	\end{tabular}
	\vspace{-2mm}
	\caption{Visual comparison on Realistic dataset~\cite{zhang2019hyperspectral} of scene 5 with corresponding PSNR. The images are from band 12 on 550 nm.}
	\vspace{-2mm}
	\label{fig:real-noisy} 
\end{figure*}
\begin{table*}[htbp]
	\centering
	\scriptsize
	\resizebox{\textwidth}{!}{
		\setlength{\tabcolsep}{2mm}
		\renewcommand{\arraystretch}{0.75}
    \begin{tabular}{c|cccccccccc}
    \bottomrule[0.8pt] 

  \rowcolor[rgb]{ .949,  .949,  .949}   \multirow{-1}{*}{Metric}        & Noisy & {BM4D  ~\cite{maggioni2012nonlocal}}  &  \makecell[c]{LLRT~\cite{chang2017hyper}}  &  \makecell[c]{NGMeet~\cite{he2019non}} &  \makecell[c]{HSID-CNN~\cite{yuan2012hyperspectral}} &  \makecell[c]{GRNet~\cite{cao2021deep}}  & \makecell[c]{QRNN3D~\cite{wei20203}} &  \makecell[c]{T3SC~\cite{bodrito2021trainable}} &\makecell[c]{MAC-Net~\cite{cao2021deep}} &\textbf{SERT (Ours)} \bigstrut\\

    \bottomrule[0.8pt]
    PSNR  &  23.26 & 29.04 & 28.26 & 28.72 & 26.44 & 25.33 & 28.12 & 28.51 & 29.20  & \textbf{29.68} \bigstrut\\
    \hline
    SSIM  & 0.7609 & 0.9471 & 0.9417 & 0.9511 & 0.8992 & 0.8381 & 0.9066 & 0.9323 & 0.9489 & \textbf{ 0.9533} \bigstrut\\
    \hline
    SAM   & 17.329 & 3.087 & 3.960 & 2.735 & 5.242 & 9.737 & 5.590 & 4.408 & 4.099 & \textbf{2.536} \bigstrut\\
    \toprule[0.8pt]
    \end{tabular}%

	}
	\vspace{-4mm}
	\caption{Average results of different methods on 15 real noisy HSIs. The PSNR is in dB, and best results are in bold.}

	\vspace{-3mm}
	\label{tab:real}%
\end{table*}%
% Table generated by Excel2LaTeX from sheet 'real'

\begin{table}[htbp]
	\centering
	\scriptsize
\resizebox{\columnwidth}{!}{
		\setlength{\tabcolsep}{1mm}
		\renewcommand{\arraystretch}{0.85}
    \begin{tabular}{c|ccccc}
		 \bottomrule[0.8pt]
	 \rowcolor[rgb]{ .949,  .949,  .949} 	
	 Metric & SwinIR~\cite{liang2021swinir} & Restormer~\cite{zamir2022restormer} & CSwin~\cite{dong2022cswin}  & TRQ3D~\cite{pang2022trq3dnet} & \textbf{SERT (Ours)} \bigstrut\\

    \bottomrule[0.8pt]
    GFLOPS &   1473.0 & 3652.8 & 1129.5 & 2135.7 & \textbf{1018.9}\bigstrut\\
    \hline
    Params (M) & 2.98  & 90.94 & 58.53 &  \textbf{0.68} & 1.91 \bigstrut\\
    \hline
    PSNR (dB)  & 40.44 & 41.07 &   42.04    & 41.66 & \textbf{42.82} \bigstrut\\
    \hline
    SSIM  & 0.9938 & 0.9945 &    0.9951   & 0.9947 & \textbf{0.9957} \bigstrut\\
    \hline
    SAM   & 2.32  & 2.05  &    2.18   & 2.21  & \textbf{1.88} \bigstrut\\
    \toprule[0.8pt]
    \end{tabular}%
}

\vspace{-4mm}
	\caption{Comparison with other Transformers under random Gaussian noise on ICVL dataset. SWinIR, Restormer, and CSWin are proposed for RBG image tasks. TRQ3D is for HSI denoising. }
	\label{tab:compare_transformer}%
	\vspace{-2mm}
\end{table}%
\noindent\textbf{Implementation Details.}
We use noise patterns in \cite{wei20203} to simulate the noisy HSIs. Specifically, the noise patterns are
 \begin{itemize}
 \vspace{-2mm}
 	\setlength{\itemsep}{0pt}
 	\setlength{\parsep}{0pt}
 	\setlength{\parskip}{0pt}
 	\item i.i.d Gaussian noise from level 10 to level 70.
 	\item Complex noise cases. Five types of complex noise are included, \emph{i.e.}, Non-i.i.d Gaussian noise, Gaussian + Stripe noise, Gaussian + Deadline noise, Gaussian + Impulse noise, and Mixture noise.
 	\vspace{-2.5mm}
 \end{itemize}

For i.i.d Gaussian noise case, we train networks with random noise levels from 10 to 70 and test them under different levels of noise. For complex noise, networks are trained with a mixture of noise and tested under each case.

For our proposed model, the learning rate is set to $1{e}{\rm-}4$ with Adam optimizer. After 50 epochs, the learning rate is divided by 10. The total epoch number is 80.  we set the basic channel $C=96$ and rank size $K=12$. The size of the rectangle of each Transformer layer is set to $[16,1], [32,2]$, and $[32,4]$ respectively. For competing methods, we use the parameter settings in the referenced works and make a great effort to reproduce the best results.
% Table generated by Excel2LaTeX from sheet 'complex'

\noindent\textbf{Quantitative Comparison.}
We show the quantitative results of Gaussian noise experiments and complex noise experiments in Tables \ref{tab:10_70_icvl} and \ref{tab:complex_icvl}. 
Among these traditional methods, NGMeet performs well on Gaussian noise cases in Table \ref{tab:10_70_icvl} and surpasses the deep learning method HSID-CNN. However, results of NGMeet and other model-based methods under complex noise cases in Table \ref{tab:complex_icvl} are much worse, showing the poor generalization ability of handcrafted priors. Our proposed method outperforms other deep learning methods by at least 0.9 dB for all noise cases. Notably, our method effectively recovers a more accurate
image from the challenging complex noisy HSIs, demonstrating its impressive ability to handle various noise.

\noindent\textbf{Visual Comparison.}
To further demonstrate the denoising performance of our
method, we show the denoised results of
different methods under random Gaussian noise and deadline noise in Figure \ref{fig:de_icvl}. In the top row, QRNN3D and QRNN3D exhibit excessive smoothness for some more complex textures. Compared to NGMeet, our method has much fewer artifacts
than other methods. In the bottom row, our method restores more texture details with less noise.

\subsection{Experiments on Real Noisy Data}
\noindent\textbf{Datasets.}
Urban dataset and Realistic dataset from \cite{zhang2021hyperspectral} are both adopted for our real data experiments. 

Urban dataset contains a image of size 307$\times$307 with 210 bands covering from $400$
to 2500 $nm$. Since there is no clean HSI, we use APEX dataset~\cite{itten2008apex} for pre-training, in which band-dependent noise levels from 0 to 55 are added to the clean HSIs. The settings are the same with \cite{bodrito2021trainable}. 
  
  For Realistic dataset\cite{zhang2021hyperspectral}, there are 59 noisy HSIs provided with paired clean HSIs. Each HSI contains 696$\times$520 pixels in spatial
  resolution with 34 bands from 400 $nm$ to 700 $nm$. We randomly select 44 HSIs from both indoor scenes and outdoor scenes. The left is used for testing.
  
\noindent\textbf{Implementation Details.}
 For Urban dataset experiment, networks are trained with their default parameter settings. The training epochs of our method is set to 100 epochs with a learning rate $1e{-4}$. 
For the Realistic dataset~\cite{zhang2019hyperspectral}, we crop overlapped 128$\times$128 spatial regions with data augmentation to train deep networks. The data augmentation settings in ~\cite{zhang2019hyperspectral} are also adopted. The training epoch is set to 1000.
\begin{table*}[htbp]
	\centering
	\scriptsize
	\resizebox{\textwidth}{!}{
		\setlength{\tabcolsep}{1.7mm}
		\renewcommand{\arraystretch}{0.75}
    \begin{tabular}{c|cccccc|cccccc}
    \bottomrule[0.8pt]
    \rowcolor[rgb]{ .949,  .949,  .949} & \multicolumn{6}{c|}{Synthetic Noise (512×512×31)} & \multicolumn{6}{c}{Real Noise (512×512×34)} \bigstrut\\
\cline{2-13}   \rowcolor[rgb]{ .949,  .949,  .949}     \multirow{-2}{*}{Metric}     & HSID-CNN & GRNet & T3SC  & QRNN3D & MAC-Net & \textbf{SERT (Ours)}  & HSID-CNN & GRNet & T3SC  & QRNN3D & MAC-Net & \textbf{SERT (Ours)} \bigstrut\\
    \hline
    PSNR (dB)  & 39.04 & 41.44 & 41.34 & 41.64 & 41.31 & 42.82 & 26.44 & 25.33 & 28.13 & 28.51 & 29.20 & 29.68\bigstrut\\
    \hline
    Params (M) & 0.40   & 44.39 & 0.83  & 0.83  & 0.43  & 1.91  & 0.40   & 44.40  & 0.83  & 0.83  & 0.43  & 1.91 \bigstrut\\
    \hline
    GFLOPS &3249.7 & 610.7 &     -  & 2513.7 &   -    &    1018.9  & 3564.2 & 611.9 &  -     &2756.9 &    -   & 1021.9 \bigstrut\\
    \hline
    Time (s) & 1.700 & 0.361 & 1.123 & 0.683 & 3.627 & 0.717 & 1.865 & 0.407 & 1.204 & 0.822 & 2.992 & 0.764 \bigstrut\\
    \toprule[0.8pt]
    \end{tabular}%
	}
	\vspace{-3.5mm}
	\caption{Comparisons of PSNR, Params, FLOPS and inference time of different deep learning methods.}
	\label{tab:comlexity}%
	\vspace{-2mm}
\end{table*}%

\noindent\textbf{Quantitative Comparison.}
Table \ref{tab:real} shows the averaged results of different methods on the Realistic dataset. Our proposed SERT significantly outperforms other HSI denoising methods by almost 0.5 dB, showing the effectiveness of our method in handling real noise. 

\noindent\textbf{Visual Comparison.}
We provide the denoising results of real noisy HSIs in Figures \ref{fig:urban_dataset} and \ref{fig:real-noisy}. Our method is superior to traditional denoising and deep learning methods in terms of both noise removal and detail retention. From  Figure \ref{fig:urban_dataset}, we can observe that Urban image is corrupted by complex noise. The stripe noise has severely affected the visual effect of image. Denoised images obtained by other methods are either over-smoothed 
or still have obvious stripe noise. Our method provides a clean
output image while preserving the textures and sharpness. For visual comparison of Realistic dataset in Figure \ref{fig:real-noisy}, the competing methods generate incorrect texture and are less effective in noise removal. And our method achieves the most promising visual result.

% Table generated by Excel2LaTeX from sheet 'Sheet3'

\begin{table*}
    \centering
    % \hspace{-0.3mm}
    \begin{subtable}[t]{0.485\linewidth}
         \centering
	\resizebox{\columnwidth}{!}{
	\setlength{\tabcolsep}{2.8mm}
		\begin{tabular}{cccccrccc}

   \bottomrule[0.8pt]
  \rowcolor[rgb]{ .949,  .949,  .949}   RA    & SE    & SS    & MU    & Params (M) & GFLOPS & PSNR (dB)  & SAM \bigstrut\\
   \hline
   $\checkmark$    &       &       &       & 1.75  &                             973.5  & 42.06 & 2.32  \bigstrut[t]\\
   $\checkmark$    &$\checkmark$    &       &       & 1.88  &                           1018.0  & 42.54 & 1.96 \\
   $\checkmark$    &$\checkmark$    &$\checkmark$    &       & 1.88  &                           1018.1  & 42.60 & 1.93 \\
   $\checkmark$    &$\checkmark$    &$\checkmark$    &$\checkmark$    & 1.91  &                           1018.9  & 42.82 & 1.88 \bigstrut[b]\\

        \toprule[0.8pt]
    \end{tabular}%
     
    }
\caption{Break-down ablation studies to verify the effectiveness of modules.}
\label{tab:component}
       
    \end{subtable}
    \hspace{3.1mm}
    \begin{subtable}[t]{0.485\linewidth}
      \centering
	\scriptsize
	\resizebox{\columnwidth}{!}{
		\setlength{\tabcolsep}{2.8mm}
		\renewcommand{\arraystretch}{0.94}
    \begin{tabular}{rcrccc}
    \bottomrule[0.8pt]
    \rowcolor[rgb]{ .949,  .949,  .949}     Method  & Params (M) & GFLOPS & PSNR (dB) & SAM \bigstrut\\
    \hline
    No SE & 1.75  &                             973.5  & 42.06 & 2.32 \bigstrut[t]\\
    Global SE & 1.91  &                           1014.8  & 42.04 & 2.22 \\
    Local SE & 1.84  &                             993.8  & 42.60 & 1.93 \\
    Non-local SE & 1.91  & 1018.9 & 42.82 & 1.88 \bigstrut[b]\\
        \toprule[0.8pt]
    \end{tabular}%

    }    \caption{Ablation to the position of spectral enhancement (SE) module.}
    \label{tab:position_of_se}
    \end{subtable}
    \vspace{-3mm}
	\caption{Component analysis of various designs on ICVL dataset under random Gaussian noise.}
	    \vspace{-4mm}
	\label{tab:ablation}
\end{table*}

\subsection{Comparison with other Transformers}
To show the effectiveness of our method in exploring spatial and spectral characteristics of HSIs, we evaluate our model with four Transformer methods in Table \ref{tab:compare_transformer}. Our model achieves the best results, implying that the proposed Transformer block is more suitable for HSI denoising. 

\noindent\textbf{Differences with existing RGB Transformers.} Existing RGB Transformer methods consider the inner long-range dependency from the spatial dimension~\cite{liu2021swin,chu2021twins} or spectral dimension~\cite{zamir2022restormer}. Our Transformer explores the joint correlation. Besides, our Transformer block utilizes the non-local similarity and low-rank property, providing a better modeling capability to explore the rich information of HSI.

\noindent\textbf{Differences with existing HSI Transformers.} TRQ3D proposed a hybrid framework that employs both Swin Transformer and 3D quasi-recurrent network for HSI denoising~\cite{pang2022trq3dnet}. With Transformer block adopted from RGB image tasks, the inner characteristic of HSI is hardly fully utilized in the proposed Transformer-based network. 

\subsection{Model Analysis}

\noindent\textbf{Model Complexity.}
In Table \ref{tab:comlexity}, we compare the average inference
time, GFLOPs as well as denoising
performance by different denoising methods on ICVL dataset and real noisy dataset~\cite{zhang2019hyperspectral}. Our method achieves the competing computation cost and inference time with better performance.

\noindent\textbf{Component Analysis.}
The results of different component designs are given in Table \ref{tab:component}. The first row presents
 Transformer with rectangle self-attention (RA) in spatial domain. Applying spectral enhancement (SE) to capture spatial-spectral information, it remarkably boosts the denoising performance by 0.42 dB improvement. The introduction of spectral shuffle (SS) also slightly improves the results, which validates the necessity of feature fusion. With memory unit (MU), the model gains 0.18 dB in PSNR, demonstrating the effectiveness of learning from a large-scale dataset to obtain representative low-rank vectors.

\noindent\textbf{Position of SE Module.}
We further place our SE module at different positions to obtain the spatial-spectral correlation. The results are shown in Table \ref{tab:position_of_se}. For global SE, the whole features of HSI is projected to one low-rank vector. Local SE stands for SE module that projected the feature inside a rectangle to one vector. Non-local SE, which is the employed design, projects several neighboring rectangles into one vector. Interestingly, global SE brings a slight decrease in performance, indicating extracting a low-rank vector from the entire HSI is inappropriate. As can
be seen that non-local SE yields the best performance. We owe it to its ability to make interactions between spatial rectangles and aggregate information of neighboring similar pixels. 

\begin{figure}[t]
	\scriptsize 
	\centering	
	
	\includegraphics[width=0.75\columnwidth]{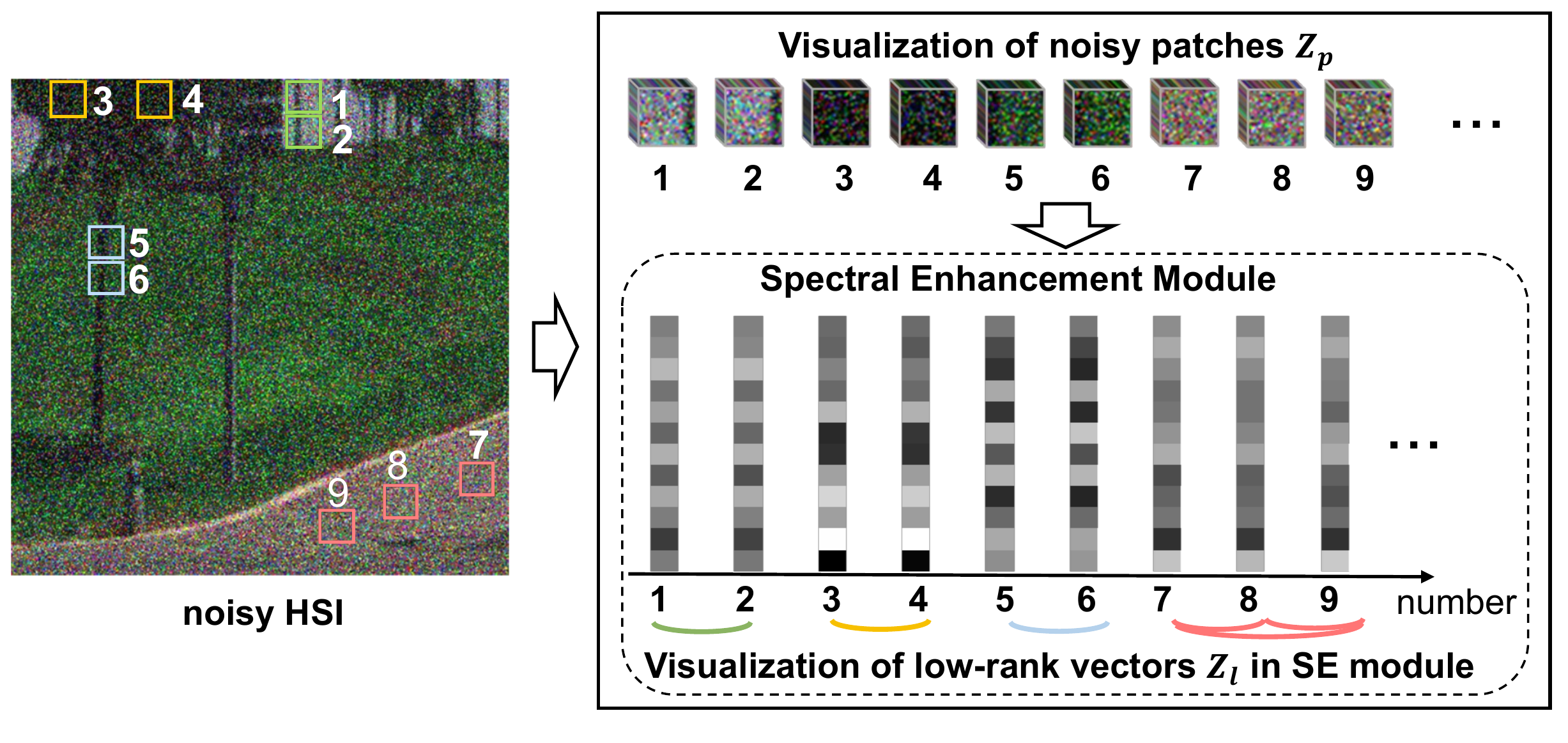}

	\vspace{-3mm}
	\caption{Visualization of low-rank vectors in SE module.}
	\label{fig:low-rank-vector}
    \vspace{-2mm}
\end{figure}

\vspace{-0.1mm}
\noindent\textbf{Visualization of Low-rank Vectors.}
To demonstrate the role of spectral enhancement module, we visualize several low-rank vectors obtained by SE module in Figure \ref{fig:low-rank-vector}. The input cubes are severely influenced by noise and it is difficult to judge the similarities between cubes visually. However, low-rank vectors extracted from these noisy cube patches by SE module show clear similarities. Since the patch 7, 8 and 9 are all from the road area, their projected low-rank vectors are more similar to each other than to other vectors. This proves the ability of SE module to extract essential information from patches and suppress noise.

\begin{figure}[t]
	\scriptsize 
	\centering	
	
	\includegraphics[width=0.8\columnwidth]{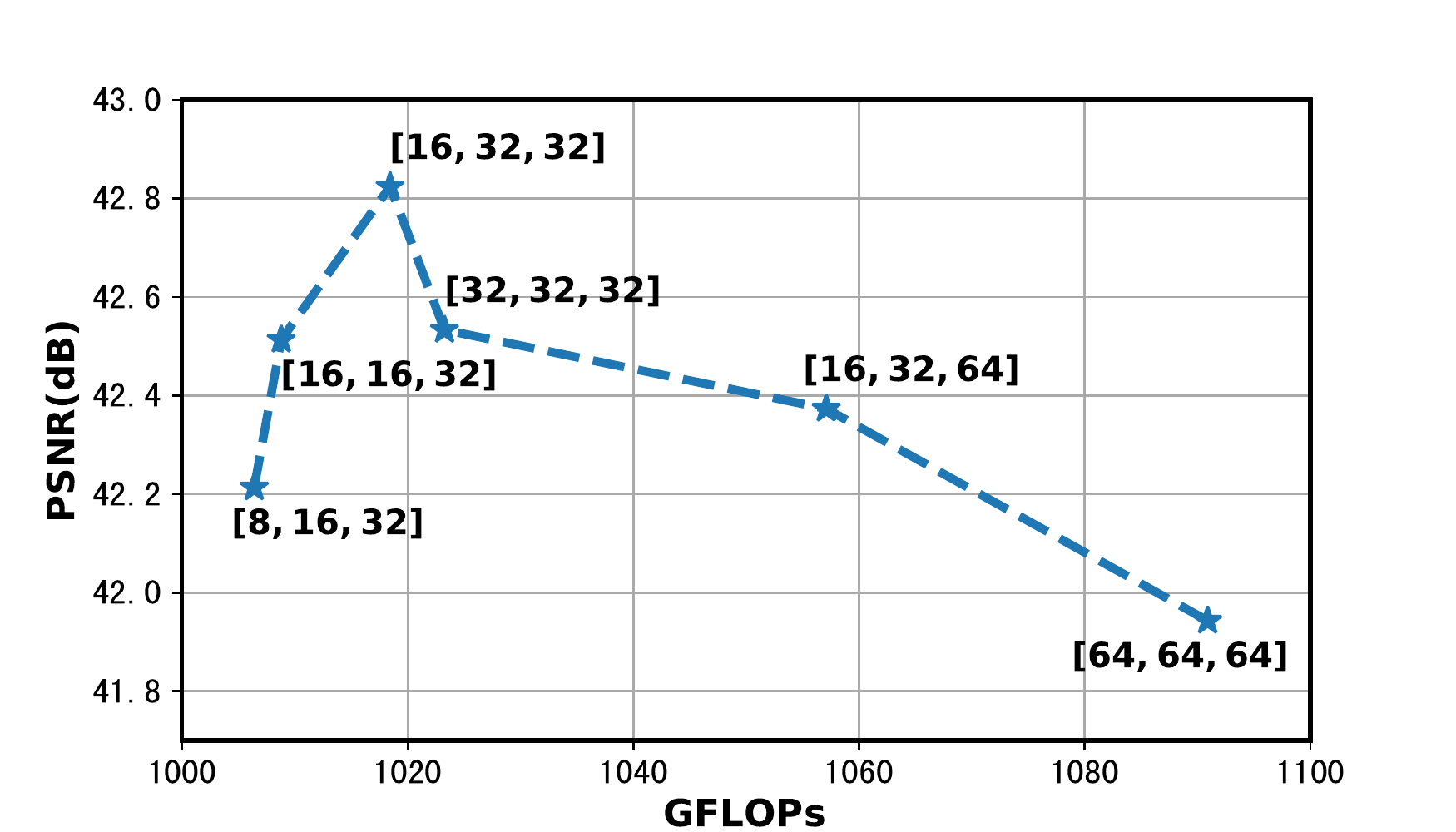}

	\vspace{-2mm}
	\caption{Different settings of rectangle's length at different layers. The widths is set to [1,2,4] for defaults.}
	\label{fig:size_rec}
    \vspace{-4.5mm}
\end{figure}

\vspace{-0.1mm}
\noindent\textbf{Parameter Analysis.}
We evaluate our proposed rectangle Transformer under different settings of rectangle size in Figure \ref{fig:size_rec}. We fix the width of rectangles and change their lengths for comparison. Since our method includes three layers of Transformer, we change the length in different layers. It can
be observed that a rectangle with longer length may not bring better performance for HSI denoising, validating the essence of our proposed rectangle self-attention in modeling non-local similarity in the spatial domain.

%------------------------------------------------------------------------
\vspace{-2mm}
\section{Conclusion}
\vspace{-1.5mm}
In this paper, we present a spectral enhanced rectangle Transformer for HSI denoising, considering the spatial non-local similarity and spectral low-rank property of HSI. We exploit the non-local similarity via multi-shape rectangle self-attention in the spatial domain with computation efficiency. Moreover, we integrate a spectral enhancement module with learnable memory unit to explore the global spectral low-rank property of HSI. The proposed spectral enhancement 
introduces interactions across spatial rectangles while maintaining informative spectral characteristics and suppressing noise. In summary, our proposed Transformer utilizes the spatial-spectral correlation to eliminate the noise. Extensive quantitative and qualitative experiments demonstrate that our method significantly outperforms  other competing methods with synthetic and real noisy HSIs. In the future, we plan to extend our method to cope with various HSI restoration tasks.

%%%%%%%%% REFERENCES
{\small
\bibliographystyle{ieee_fullname}
\bibliography{egbib}
}

\end{document}